\documentclass[10pt]{article}
\usepackage[utf8]{inputenc} 
\usepackage[T1]{fontenc}    
\usepackage{url}            
\usepackage{booktabs}       
\usepackage{amsfonts}       
\usepackage{nicefrac}       
\usepackage{microtype}      

\usepackage[round]{natbib}
\usepackage{fullpage} 

\usepackage{latexsym,amsbsy,amssymb,amsmath,color, url,booktabs, multirow, xspace, graphicx }
\usepackage{xspace}

\newcommand{\commentOut}[1]{} 

\newcommand{\mat}[1]{\mathbf{#1}}
\renewcommand{\vec}[1]{ \mathbf{#1} } 
\newcommand{\vecS}[1]{\boldsymbol{ #1 }  } 

\newcommand{\C}{\mat{C}}

\newcommand{\I}{\mat{I}}
\newcommand{\K}{\mat{K}}

\newcommand{\X}{\mat{X}}

\newcommand{\Z}{\mat{Z}}

\newcommand{\calD}{\mathcal{D}} 
\newcommand{\calL}{\mathcal{L}}

\newcommand{\calX}{\mathcal{X}}

\newcommand{\calY}{\mathcal{Y}}
\newcommand{\calO}{\mathcal{O}}

\providecommand{\abs}[1]{\lvert#1\rvert}

\newcommand{\expectation}{\mathbb{E}}
\newcommand{\Eb}[1]{\left\langle #1 \right\rangle} 

\newcommand{\Normal}{\mathcal{N}}

\newcommand{\kernel}{\kappa}

\newcommand{\gradient}{\nabla}

\newcommand{\trace}{\mbox{ \rm tr }}
\renewcommand{\det}[1]{\left\lvert#1\right\rvert}
\newcommand{\defeq}{\stackrel{\text{\tiny def}}{=}}

\newcommand{\mth}{\mathrm{th}}

\newcommand{\bigO}{\calO}

\newcommand{\kl}[2]{\mathrm{KL}(#1 \lVert #2)}

\newcommand{\name}[1]{{\sc #1}\xspace}
\newcommand{\gp}{\name{gp}}
\newcommand{\church}{\name{church}}
\newcommand{\stan}{\name{stan}}
\newcommand{\mcmc}{\name{mcmc}}
\newcommand{\vi}{\name{vi}}

\newcommand{\crf}{\name{crf}}
\newcommand{\ess}{\name{ess}}
\newcommand{\svm}{\name{svm}}
\newcommand{\gpess}{\name{gp-ess}}
\newcommand{\gpvarb}{\name{gp-var-b}}
\newcommand{\gpvars}{\name{gp-var-s}}
\newcommand{\gppseudo}{\name{gp-var-p}}
\newcommand{\gpstruct}{\name{gpstruct}}
\newcommand{\sgd}{\name{sgd}}
\newcommand{\saga}{\name{saga}}
\newcommand{\basenp}{\name{base np}}
\newcommand{\chunking}{\name{chunking}}
\newcommand{\segmentation}{\name{segmentation}}
\newcommand{\japanesene}{\name{japanese ne}}
\newcommand{\nlp}{\name{nlp}}

\newcommand{\hyper}{\vecS{\theta}}
\newcommand{\varparam}{\llambda}

\newcommand{\mydot}{\cdot}
\newcommand{\iid}{i.i.d\xspace}
\newcommand{\indicator}[1]{\mathbb{I}[#1]}
\newcommand{\Ajn}{\mat{A}_{jn}}
\newcommand{\tildeKjn}{\priorcov^{(n)}}
\newcommand{\Xn}{\mat{X}_n}

\newcommand{\funnj}{\f_{\un n j}}
\newcommand{\bkjn}{\qfmean{kj(n)}}
\newcommand{\Sigmakjn}{\qfcov{kj(n)}}

\newcommand{\un}{\text{u}}
\newcommand{\fun}{\f_{\un}}
\newcommand{\funn}{\fun^{(n)}}
\newcommand{\funnki}{{\fun}_{n \mydot}^{(k,i)}}
\newcommand{\funnotn}{\fun^{\backslash n}}
\newcommand{\funj}{\f_{\un \cdot j}}

\newcommand{\fbinni}{\fbin^{(i)}}
\newcommand{\bin}{\text{bin}}

\newcommand{\tn}{T_{n}} 
\newcommand{\V}{{\mathcal V}} %
\newcommand{\Vsize}{\abs{\V}} %
\newcommand{\fbin}{\f_{\bin}}

\newcommand{\Kbin}{\K_{\bin}}

\newcommand{\Kj}{\mat{K}_j}
\newcommand{\mkj}{\postmean{kj}}
\newcommand{\Skj}{\postcov{kj}}
\newcommand{\mbin}{\postmean{\bin}}
\newcommand{\Sbin}{\postcov{\bin}}

\newcommand{\qun}{q(\fun)}
\newcommand{\qbin}{q(\fbin)}
\newcommand{\elltermkn}{\ellterm^{(k,n)}} 

\newcommand{\lambdaku}{\llambda_{k}^{\un}}
\newcommand{\lambdabin}{\llambda_{\bin}}
\newcommand{\lambdaun}{\llambda_{\un}}

\newcommand{\df}{\text{d}\f}
\newcommand{\du}{\text{d}\u}

\newcommand{\dfunn}{\text{d} \funn}
\newcommand{\dfbin}{\text{d} \fbin}

\newcommand{\qkun}{q_k(\fun)}
\newcommand{\qkunn}{q_{k(n)}(\funn)}

\newcommand{\priormean}{\tilde{\vecS{\mu}}_j} 
\newcommand{\priorcov}{\widetilde{\K}_j}
\newcommand{\postmean}[1]{\vec{m}_{#1}}
\newcommand{\postcov}[1]{\mat{S}_{#1}}

\newcommand{\bs}{\boldsymbol}
\newcommand{\llambda}{\bs{\lambda}}

\newcommand{\kernelj}{\kernel_j}
\newcommand{\Kzzall}{\K_{zz}} 
\newcommand{\Kzzallinv}{\K^{-1}_{zz}} 
\newcommand{\Kzz}{\kernel(\Zj, \Zj)}
\newcommand{\Kxx}{\kernel_j(\X,\X)}
\newcommand{\Kxz}{\kernel(\X,\Z_j)}
\newcommand{\Kzx}{\kernel(\Z_j, \X)}
\newcommand{\kzn}{\kernel(\Z_j, \X_n)}
\newcommand{\knz}{\kernel(\X_n, \Z_j)}
\newcommand{\knn}{\kernelj(\X_n, \X_n)}
\newcommand{\Kzzinv}{\kernel(\Zj, \Zj)^{-1}}

\newcommand{\nseq}{ {N_{\text{seq}}} } 
\newcommand{\n}{N} 
\newcommand{\m}{M} 

\renewcommand{\dim}{D} 
\renewcommand{\k}{K} 

\newcommand{\indexdata}[2]{#1^{(#2)}}
\newcommand{\indexclique}[2]{#1_{#2}}
\newcommand{\inputset}{\calX}
\newcommand{\outputset}{\calY}
\renewcommand{\X}{\mat{X}}
\newcommand{\y}{\vec{y}} 
\newcommand{\x}{\X}

\newcommand{\xn}{\indexdata{\x}{n}}
\newcommand{\yn}{\indexdata{\y}{n}} 
\newcommand{\xstar}{\indexdata{\x}{\star}}
\newcommand{\ystar}{\indexdata{\y}{\star}} 

\newcommand{\xc}{\indexclique{\x}{c}}
\newcommand{\yc}{\indexclique{\y}{c}}
\newcommand{\yprime}{\y^\prime}
\newcommand{\ycprime}{\yc^\prime}
\newcommand{\f}{\vec{f}} 
\newcommand{\fn}{\f_{n \mydot}} 
\newcommand{\fj}{\f_{\mydot j}} 
\renewcommand{\u}{\vec{u}} 
\newcommand{\uj}{\vec{u}_{\mydot j} } 
\renewcommand{\K}{\mat{K}}
\newcommand{\Zj}{\mat{Z}_j}
\newcommand{\Aj}{\mat{A}_j}
\newcommand{\data}{\calD}

\newcommand{\elbo}{\calL_{\text{elbo}}}
\newcommand{\ellterm}{\calL_{\text{ell}}}
\newcommand{\klterm}{\calL_{\text{kl}}}
\newcommand{\enterm}{\calL_{\text{ent}}}
\newcommand{\crossterm}{\calL_{\text{cross}}}
\newcommand{\elltermhat}{\widehat{\calL}_{\text{ell}}}  
\newcommand{\entermhat}{\hat{\calL}_{\text{ent}}}
\newcommand{\klun}{\klterm^{\un}}
\newcommand{\klbin}{\klterm^{\bin}}

\newcommand{\qu}{q(\u)}
\newcommand{\qf}{q(\f)}

\newcommand{\qfmean}[1]{\vec{b}_{#1}}        
\newcommand{\qfcov}[1]{\mat{\Sigma}_{#1}}        

\newcommand{\grad}{\gradient}

\newcommand{\nkl}{\Normal_{k \ell}}
\newcommand{\zk}{z_{k}}
\newcommand{\zl}{z_{\ell}}
\newcommand{\Ckl}{\C_{kl}}
\newcommand{\pil}{\pi_{\ell}}

\newcommand*{\QEDA}{\hfill\ensuremath{\blacksquare}}%

\title{Gray-box inference for\\ structured Gaussian process models}

\author{Pietro Galliani$^1$,~Amir Dezfouli$^2$,~Edwin V. Bonilla$^3$, and~Novi~Quadrianto$^1$ \vspace{0.5cm}\\
$^1$SMiLe CLiNiC, University of Sussex, Brighton, UK\\
$^2$Data61, CSIRO, Sydney, Australia\\
$^3$The University of New South Wales, Sydney, Australia\\
}

\begin{document}

\maketitle

\begin{abstract} 
We develop an automated variational inference method for 
Bayesian structured prediction problems with Gaussian process (\gp) priors 
and linear-chain likelihoods. 
Our approach does not  need to know the details of the structured likelihood model 
and can scale up to a large 
 number of observations. 
Furthermore, we show that the required expected likelihood term and its 
gradients in the variational objective (ELBO) can be 
estimated efficiently by using expectations 
over very low-dimensional Gaussian distributions. Optimization of the 
 ELBO is 
fully parallelizable over sequences and  amenable to stochastic optimization,
which we use along with control variate techniques and state-of-the-art 
incremental optimization to  make our framework useful in practice. 
Results on a set of natural language processing tasks show that our method can be as good as (and 
sometimes better than)  hard-coded approaches including \svm-struct and \crf{s}, and overcomes the 
scalability limitations of previous inference algorithms based on 
sampling. 
Overall, this is a fundamental step to developing automated inference methods 
for Bayesian structured prediction.
\end{abstract}

\section{Introduction}
Developing automated inference methods for complex probabilistic models 
has become arguably one of the most exciting areas of research in machine learning, 
with notable examples in the probabilistic programming community given by
\stan \citep{hoffman-gelman-jmlr-2014} and  \church \citep{goodman-et-al-2008}.
One of the main challenges for these types of approaches is to formulate expressive
probabilistic models and develop generic yet efficient inference methods for them. 
From a variational inference perspective, one particular approach that has 
addressed such a challenge is the black-box variational inference framework 
of  \citet{ranganath-et-al-aistats-2014}. 

While the works of \citet{hoffman-gelman-jmlr-2014}  and \citet{ranganath-et-al-aistats-2014}
have been  successful with a wide range of priors and likelihoods, 
their direct application to models with Gaussian process (\gp) priors is cumbersome,
mainly due to the large number of highly coupled latent variables in such models.  
In this regard, very recent work  has investigated automated inference methods 
for general likelihood models when the prior is given by a sparse Gaussian process 
\citep{hensman-et-al-nips-2015, dezfouli-bonilla-nips-2015}. 
While these advances have opened up opportunities for applying \gp-based models 
well beyond regression and classification settings, they have focused 
on models with \iid observations and, therefore,  are 
unsuitable for addressing the more challenging task of \emph{structured prediction}.
 
Structured prediction refers to the problem where there are interdependencies 
between the outputs and it is necessary to model these dependencies explicitly.  
Common examples are found in natural 
language processing (\nlp) tasks, computer vision 
and bioinformatics. By definition, observation models in these problems are not \iid 
and  standard learning frameworks have been extended to consider the 
constraints imposed by structured prediction tasks.  Popular 
structured prediction frameworks are  
conditional random fields \citep[\crf{s};][]{lafferty-et-al-icml-2001}, 
maximum margin Markov networks \citep{taskar-et-al-nips-2004}  and 
structured support vector machines \citep[\svm-struct,][]{tsochantaridis-et-al-jmlr-2005}.

From a non-parametric Bayesian modeling perspective, in general, and from a \gp modeling 
perspective, in particular, structured prediction problems present incredibly
hard inference challenges because of  the rapid explosion of the number of latent variables
with the size of the problem. Furthermore, structured likelihood functions are usually 
very expensive to compute. In an attempt to build non-parametric Bayesian approaches 
to structured prediction, \citet{bratieres-et-al-tpami-2015} have proposed 
a  framework based on a \crf-type modeling approach with 
\gp{s},  and use elliptical slice sampling 
\citep[\ess;][]{murray-eyt-al-aistats-2010} as part of their inference method.
 Unfortunately, although their method can be applied to  linear chain 
 structures in a generic way without considering the details of the likelihood model,
 it is not scalable as it involves sampling from the full \gp prior. 
 
 In this paper we present an approach for automated  inference 
 in structured \gp models with linear chain likelihoods that builds upon 
  the structured \gp model of \citet{bratieres-et-al-tpami-2015} and the 
  sparse variational frameworks of \citet{hensman-et-al-nips-2015}
  and \citet{dezfouli-bonilla-nips-2015}.  
  In particular, we show that the model of \citet{bratieres-et-al-tpami-2015} can 
  be mapped onto a generalization of the automated inference framework
  of \citet{dezfouli-bonilla-nips-2015}.
  Unlike the work of \citet{bratieres-et-al-tpami-2015},
  by introducing sparse \gp priors in structured prediction models, our approach
  is  scalable to a large number of observations. More importantly, 
  this approach is also generic in that it does not need to know the details 
  of  the likelihood model 
  in order to carry out posterior inference.
  Finally, we show that our inference method is statistically efficient as, despite 
  having a Gaussian process prior over a large number of latent functions,  
  it only requires expectations over low-dimensional Gaussian distributions in order 
  to carry out posterior approximation.

Our experiments on a   set of \nlp tasks, including noun phrase 
identification, chunking, segmentation, and  named entity recognition,
show that our method can be as good as (and 
sometimes better than)  hard-coded approaches including \svm-struct and \crf{s}, and overcomes the 
scalability limitations of previous inference algorithms based on 
sampling. 

We refer to our approach as ``gray-box'' inference 
since, in principle, for general structured prediction problems it may require some human 
intervention. Nevertheless, when applied 
to fixed structures, our proposed inference method   
is entirely “black box”.

\section{Gaussian process models for structured prediction}
Here we are interested in structured prediction problems where we observe input-output pairs
$\data = \{ \xn, \yn \}_{n=1}^\nseq$, where $\nseq$ is the total number of observations,
$\xn \in \inputset$ is a descriptor of observation $n$ and $\yn \in \outputset$ is a structured  object such as 
a sequence, a tree or a grid that reflects the interdependences between its individual constituents.  Our goal 
is that of, given a new input descriptor $\xstar$, predicting its corresponding structured labels $\ystar$, and 
more generally, a distribution over these labels.  

A fairly general approach to address this problem with 
Gaussian process (\gp) priors was proposed by   \citet{bratieres-et-al-tpami-2015} based on 
\crf-type models, where the distribution of the output given the input is defined in terms of cliques, 
i.e.~sets of fully connected nodes. Such a distribution is given by:
\begin{align}
	\label{eq:struct-softmax}
	p(\y | \x, \f) = \frac{\exp\left(\sum_c f(c, \xc, \yc) \right)}{\sum_{\yprime \in \outputset} \exp \left( \sum_c f(c, \xc, \ycprime) \right) }
	\text{,}
\end{align} 
where $\xc$ and  $\yc$ are tuples of nodes belonging to clique $c$;  $f(c, \xc, \yc)$ 
is their corresponding latent variable; and $\f$ is the collection of all these latent variables,
which are assumed to be drawn from a zero-mean \gp prior with covariance function $\kernel(\mydot, \mydot; \hyper)$, 
with $\hyper$ being the hyperparameters.  
It is clear that such a model is a generalization of vanilla \crf{s} where the potentials
 are draws from a \gp instead of 
being linear functions of the features.  %
\subsection{Linear chain structures \label{sec:linear-chain}}
In this paper we focus on linear chain structures where both the input and the output 
corresponding to datapoint $n$ are linear chains of length $\tn$, whose corresponding constituents 
stem from a common set.  In other words, $\xn$ is a $\tn \times \dim$ matrix of feature descriptors and 
$\yn$ is a sequence of $\tn$ labels drawn from the same vocabulary  $\V$. In this case, in order to completely define 
the prior over the clique-dependent latent functions in Equation \eqref{eq:struct-softmax}, it is necessary to specify 
 covariance functions over the cliques. To this end, \citet{bratieres-et-al-tpami-2015} propose a kernel 
that is non-zero only  when two cliques are of the same type, i.e.~both  are unary cliques or both  are pairwise cliques. Furthermore, these kernels are defined as:
\begin{align}
	\kernel_{\un}((t, \vec{x}_t, y_t),(t^\prime, \vec{x}_t^\prime,y_{t^\prime}))
	& = \indicator{y_t = y_{t^\prime}} \kernel(\vec{x}_t, \vec{x}_{t}^{\prime}) \\
	\kernel_{\bin}((y_t, y_{t+1}),(y_{t^\prime}, y_{t^\prime +1}))
	& = \indicator{y_t = y_{t^\prime} \wedge  y_{t+1} = y_{t^\prime +1}  } \text{,}
\end{align}
where $\kernel_{\un}$ is the covariance on unary functions and 
$\kernel_{\bin}$ is the covariance on pairwise functions.  With a suitable ordering of these 
latent functions, we obtain a posterior covariance matrix that is block-diagonal, 
with the first $\Vsize$ blocks corresponding to the unary covariances, each of 
size $\tn$; and the last block, corresponding to the pairwise covariances, being a 
diagonal (identity) matrix of size $\Vsize^{2}$, where $\Vsize$ denotes the vocabulary size.

 To carry out inference in this model,  
 \citet{bratieres-et-al-tpami-2015} propose a sampling scheme based on 
 elliptical slice sampling \citep[\ess;][]{murray-eyt-al-aistats-2010}. In the following section,  
 we show an equivalent  formulation of this model that leverages the general 
 class of models with \iid likelihoods presented by  \citet{nguyen-bonilla-nips-2014}. 
 Understanding structured \gp models from such a perspective will allow us to generalize 
 the results of   \citet{nguyen-bonilla-nips-2014,dezfouli-bonilla-nips-2015} in order to develop 
an automated variational inference framework. The advantages of such a framework are 
that of (i) dealing with generic likelihood models; and (ii) enabling stochastic optimization 
techniques for scalability to large datasets.

\section{Full Gaussian process priors and automated inference \label{sec:full-gp}}
%
%

\citet{nguyen-bonilla-nips-2014} developed an automated variational
inference framework for a class models with Gaussian process priors 
and generic \iid likelihoods. Although such an approach is an important step 
towards black-box inference with \gp priors, assuming \iid observations 
is, by definition,  unsuitable for structured models.  
 
 One way to generalize such an approach to structured models of the types 
 described in \S \ref{sec:linear-chain} is to differentiate between \gp priors over 
 latent functions on unary nodes and  \gp priors over 
 latent functions  over pairwise nodes. More importantly,
 rather than considering \iid likelihoods over all observations, we assume likelihoods that factorize  
 over sequences, while allowing for statistical dependences within a sequence. 
 Therefore, our prior model for linear chain structures is given by:
\begin{align}
\label{eq:full-prior}
p(\f) = p(\fun) p(\fbin)  = \left( \prod_{j=1}^{\Vsize} \Normal(\funj; \vec{0}, \Kj) \right) \Normal(\fbin; \vec{0}, \Kbin) \text{,}
\end{align}
where $\f$ is the vector of all latent function values of unary nodes $\fun$ and the  function values of pairwise  nodes $\fbin$. 
 Accordingly, $\funj$ is the vector of unary  functions of latent process $j$, corresponding 
 to the $j\mth$ label in the vocabulary, which is drawn from a zero-mean \gp with 
 covariance function $\kernelj(\mydot, \mydot; \hyper_j)$. This covariance function, when evaluated 
 at all the input pairs in $\{\xn\}$, induces the 
 $\n \times \n$ covariance matrix $\Kj$, where $\n = \sum_{n=1}^\nseq \tn$  is the total 
 number of observations. 
 Similarly, $\fbin$ is a zero-mean $\Vsize^2$-dimensional Gaussian random variable with 
  covariance matrix given by $\Kbin$.  We note here that while the unary functions 
are draws from a \gp indexed by $\X$, the distribution over pairwise functions is a finite 
Gaussian (not indexed by $\X$).

Given the latent function values, our conditional likelihood is defined by: 
\begin{align}
\label{eq:likelihood}
p(\y | \f) = \prod_{n=1}^\nseq p(\yn | \fn) \text{,}
\end{align}
where, omitting the dependency on the input $\x$ for simplicity, each individual 
conditional likelihood term is computed using a valid likelihood function for sequential 
data such as that defined by the structured softmax function in Equation \eqref{eq:struct-softmax};
$\yn$ denotes the labels of sequence $\yn$; and 
$\fn$ is the corresponding vector of latent (unaries and pairwise) function values.
We now have all the necessary definitions to state our first result.

\newtheorem{theorem1}{Theorem}
\begin{theorem1}
The model class defined by the prior in Equation \eqref{eq:full-prior} and 
the likelihood in Equation \eqref{eq:likelihood} contains the structured 
\gp model proposed by \citet{bratieres-et-al-tpami-2015}. 
\end{theorem1}
The proof of this is trivial and can be done by (i) setting all the 
covariance functions of the unary latent process ($\kernel_j$) to be the same; 
(ii) making  $\Kbin = \I$; and (iii) using the structured softmax function in 
Equation \eqref{eq:struct-softmax} as each of the individual terms $p(\yn | \fn)$
in Equation $\eqref{eq:likelihood}$. This yields exactly the same model as 
specified  by \citet{bratieres-et-al-tpami-2015},  with 
prior covariance matrix with block-diagonal structure described 
in \S \ref{sec:linear-chain} above. 
\QEDA

The practical consequences of the above theorem is that we can now leverage 
the results of \citet{nguyen-bonilla-nips-2014} in order to develop a variational inference (\vi) framework
for structured \gp models that can be carried out without knowing the details of the 
conditional likelihood. Furthermore, as we shall see in the next section, in order to deal 
with the intractable nonlinear expectations inherent to \vi , the proposed method only requires expectations 
over low-dimensional Gaussian distributions.
\subsection{Automated variational inference}
In this section we develop a method for estimating the posterior over the latent 
functions given the prior and likelihood models defined in Equations \eqref{eq:full-prior}
and \eqref{eq:likelihood}. Since the posterior is analytically intractable and the 
prior involves a large number of coupled latent variables, we resort to approximations 
given by variational inference \citep[\vi; ][]{jordan-etl-al-book-1998}. To this end, we start by 
defining our variational approximate posterior distribution:
\begin{align}
\label{eq:q-full-init}
  \qf &= \qun \qbin 
   \text{,} \quad \text{ with } \\
  \qun &= \sum_{k=1}^K \pi_k q_k(\fun | \qfmean{k}, \qfcov{k}) = \sum_{k=1}^K \pi_k \prod_{j=1}^{\Vsize} \Normal(\funj; \qfmean{kj}, \qfcov{kj}) 
  \quad \text{ and } \\
\label{eq:q-full-end}
  \qbin & = \Normal(\fbin; \mbin, \Sbin)  \text{,}
\end{align}
where $\qun$ and $\qbin$ are the approximate posteriors over 
the unary and pairwise nodes respectively; 
each $q_k(\funj)= \Normal(\funj; \qfmean{kj}, \qfcov{kj})$ is a $\n$-dimensional full Gaussian distribution; and 
 $\qbin$ is a $\Vsize^2$-dimensional Gaussian.

In order to estimate the parameters of the above distribution, variational inference entails the 
optimization of the so-called evidence lower bound ($\elbo$), which can be shown to be a lower bound 
of the true marginal likelihood, and  is composed of a KL-divergence term ($\klterm$), between the 
approximate posterior and the prior,  and an expected log likelihood term ($\ellterm$):
\begin{equation}
\label{eq:elbo-full}
\elbo = - \kl{\qf}{p(\f)} + \Eb{\log p(\y | \f) }_{\qf} \text{,}
\end{equation}
where the angular bracket notation $\Eb{\mydot}_{q}$ indicates an expectation over the distribution $q$.  Although 
the approximate posterior is an $\n$-dimensional distribution, 
the expected log likelihood term can be estimated efficiently using expectations over much lower-dimensional
Gaussians.
\begin{theorem1}
\label{th:efficient-full}
For the structured \gp model  defined in Equations \eqref{eq:full-prior} and 
\eqref{eq:likelihood}, 
the expected log likelihood over the variational distribution defined in 
Equations \eqref{eq:q-full-init} to 
\eqref{eq:q-full-end} 
 and its gradients can
be estimated using expectations over $\tn$-dimensional Gaussians and 
$\Vsize^2$-dimensional Gaussians, where 
$\tn$ is the length of each sequence and $\Vsize$ is the vocabulary size.
\end{theorem1}
The proof is constructive and can be found in the supplementary material. Here we 
state the final result on how to compute these estimates:
\begin{align}
\label{eq:estimates-full-begin}
\ellterm &= \sum_{n=1}^{\nseq} \sum_{k=1}^{K} \pi_k  \Eb{\log p(\yn | \fn)}_{\qkunn \qbin }  \text{,} \\
\grad_{\lambdaku}  \elltermkn &= \Eb{\grad_{\lambdaku} \log \qkunn \log p(\yn | \fn)}_{\qkunn \qbin } \text{,} \\
\label{eq:estimates-full-end}
 \grad_{\lambdabin}  \elltermkn &= \Eb{\grad_{\lambdabin} \log \qbin \log p(\yn | \fn)}_{\qkunn \qbin } \text{,}
\end{align}
where $ \qkunn$ is a $(\tn \times \Vsize)$-dimensional Gaussian with block-diagonal covariance $ \qfcov{k(n)}$, each 
block of size $\tn \times \tn$. Therefore, we can estimate the above term by sampling from 
$\tn$-dimensional Gaussians independently. Furthermore, $\qbin$ is a $\Vsize^2$-dimensional Gaussian, which 
can also be sampled independently. In practice, we can 
assume that the covariance of $\qbin$ is diagonal and we only sample from univariate Gaussians for the 
pairwise functions.

It is important to emphasize the practical consequences of Theorem \ref{th:efficient-full}. Although we 
have a fully correlated prior and a fully correlated approximate posterior over  $\n = \sum_{n=1}^{\nseq} \tn$ 
unary function values, yielding full $\n$-dimensional covariances, we have shown that for these classes of models 
we can estimate $\ellterm$ by only using expectations  over $\tn$-dimensional Gaussians. We refer 
to this result as that of \emph{statistical efficiency} of the inference algorithm.

Nevertheless, even when having only one latent function and 
using a single Gaussian approximation ($K=1$), optimization 
of the $\elbo$ in Equation \eqref{eq:elbo-full} is completely impractical for any realistic dataset concerned 
with structured prediction problems, 
due to its high memory requirements $\bigO(\n^2)$ and time complexity $\bigO(\n^3)$. 
In the following section we 
will use a sparse \gp approach within our variational framework in order to develop a practical algorithm 
for structured prediction.

\section{Sparse Approximation}
In this section we describe a scalable approach to inference in the 
structured \gp model defined in \S \ref{sec:full-gp} by  introducing the so-called sparse \gp approximations
\citep{quinonero2005unifying} into our variational framework.  Variational approaches to sparse \gp 
models were  developed by \citet{titsias-aistats-2009} for  Gaussian \iid likelihoods,  then 
made scalable to large datasets and generalized to non-Gaussian (\iid) likelihoods by
\citet{hensman-et-al-aistats-2015, hensman-et-al-nips-2015,dezfouli-bonilla-nips-2015}.
The main idea of such approaches is to introduce a set of $\m$ \emph{inducing variables} 
$\{ \uj \}_{j=1}^\m$ for each latent process,
which  lie in the same space as  $\{ \fj \}$ and are drawn from the same \gp prior.
These inducing variables are the latent function values of their corresponding set of \emph{inducing inputs} 
$\{ \Zj \} $. Subsequently, we redefine our prior in terms of these inducing inputs/variables.

In our structured \gp model, only the unary latent functions are drawn from \gp{s} indexed by $\x$. Hence 
we assume a \gp prior over the inducing variables and a conditional prior  over the unary latent functions,
 which both factorize  over the latent processes, yielding the joint distribution over 
 unary functions, pairwise functions and inducing variables given by:
 \begin{equation}
 \label{eq:prior-sparse}
  p(\f, \u) = p(\u) p(\fun | \u)  p(\fbin)   \text{, with }    
  p(\fun  | \u)  = \prod_{j=1}^{\Vsize} \Normal(\funj; \priormean, \priorcov) \text{ and }   
  p(\u) = \prod_{j=1}^{\Vsize} p(\uj)  \text{,}
 \end{equation}
with the prior over the pairwise functions defined as before, i.e.~$p(\fbin) = \Normal(\fbin; \vec{0}, \Kbin)$, 
and the means and covariances of the conditional  distributions over the unary functions are given by:
\begin{align}
  \priormean &=\Aj \uj  \text{ and }   \priorcov = \Kxx - \Aj \Kzx \text{, with }  \Aj  = \Kxz \Kzzinv  \text{.} 
\end{align}
By keeping an explicit representation of the inducing variables, our goal is to estimate the joint 
posterior over the unary functions, pairwise functions and inducing variables given the observed data.
To this end,  we assume that our variational approximate  posterior is given by:
\begin{equation}
  \label{eq:posterior-sparse}
  q(\f, \u | \llambda) = p(\fun | \u) q(\u | \lambdaun) q(\fbin | \lambdabin) \text{,}
\end{equation}
where $\llambda = \{ \lambdaun, \lambdabin\}$ are the variational parameters;
 $p(\fun | \u)$ is defined in Equation \eqref{eq:prior-sparse};
 $q(\fbin | \lambdabin)$ is defined as in Equation \eqref{eq:q-full-end},  i.e.~a Gaussian 
 with parameters $\lambdabin = \{\mbin, \Sbin \}$;
  and
\begin{equation}
q(\u | \lambdaun) = \sum_{k=1}^{\k} \pi_k q_k(\u | \postmean{k}, \postcov{k}) = \sum_{k=1}^{\k} \pi_k \prod_{j=1}^{\Vsize}  \Normal(\uj; \postmean{kj}, \postcov{kj}) \text{,}
\end{equation}
with  $\lambdaun = \{\pi_k, \postmean{k}, \postcov{k} \}$  and 
$\postmean{kj}, \postcov{kj}$ denoting the posterior mean and covariance
of the inducing variables corresponding to mixture component $k$ and latent  function $j$. 
\subsection{Evidence lower bound}
The KL term in the evidence lower bound  now considers a KL divergence between 
the joint approximate posterior  in Equation \eqref{eq:posterior-sparse} and the joint prior 
in Equation \eqref{eq:prior-sparse}.  Because of the structure of the approximate posterior, 
it is easy to show that the term $ p(\fun | \u)$ vanishes from the KL, yielding 
an objective function that is composed of a KL between the distributions over the inducing variables; 
a KL between the distributions over the pairwise functions, and the expected log likelihood 
over the joint approximate posterior:
\begin{align}
	\label{eq:elbo-sparse}
	\elbo(\llambda) = - \kl{q(\u)}{p(\u)} - \kl{q(\fbin)}{p(\fbin)}   + \Eb{\sum_{n=1}^\nseq \log p(\yn | \fn)}_{  q(\f, \u | \llambda) } \text{,}
\end{align}
where $\kl{q(\fbin)}{p(\fbin)}$ is a straightforward KL divergence between two Gaussians and 
$\kl{q(\u)}{p(\u)}$ is a KL divergence between a Mixture-of-Gaussians and a Gaussian, which we bound 
using Jensen's inequality. The expressions for these terms are given in the supplementary material.
 
Let us now consider the expected log likelihood term in Equation \eqref{eq:elbo-sparse}, which is an expectation 
of the conditional likelihood over the joint posterior $q(\f,\u | \llambda)$. The following result tells us that, as in 
the full (non-sparse) case, these expectations can still be estimated efficiently by using expectations over low-dimensional 
Gaussians. 
\begin{theorem1}
\label{th:efficient-sparse}
The expected log likelihood term in Equation \eqref{eq:elbo-sparse}, with 
a generic structured conditional likelihood   $p(\yn | \fn)$ and  
variational distribution $q(\f,\u | \llambda)$ defined in Equation 
 \eqref{eq:prior-sparse},  and its gradients can
be estimated using expectations over $\tn$-dimensional Gaussians and 
$\Vsize^2$-dimensional Gaussians, where 
$\tn$ is the length of each sequence and $\Vsize$ is the vocabulary size.
\end{theorem1}
As in the full (non-sparse) case, the proof is constructive and can be found in the 
supplementary material. 
This means that,  in the sparse 
case,  
the expected log likelihood and its gradients can also be  computed using  Equations 
\eqref{eq:estimates-full-begin} to  
\eqref{eq:estimates-full-end}, where the mean and covariances 
of each $\qkunn$ are determined by the means and covariances of the 
posterior over the inducing variables. 
Thus, as before,  $\qkunn$ is a ($\tn \times \Vsize$)-dimensional Gaussian with block-diagonal 
structure, where each of the $j=1, \dots, \Vsize$ blocks has mean and covariance given by:
\begin{eqnarray}
\label{eq:bkjn-skjn}
  \bkjn = \Ajn \mkj \text{,} & 
    \Sigmakjn  = \tildeKjn + \Ajn \Skj \Ajn^T \text{, where} \\
  \Ajn \defeq \knz \Kzzinv \quad \text{ and } &
  \tildeKjn \defeq \knn - \Ajn \kzn \text{,} 
\end{eqnarray}
where, as mentioned in \S \ref{sec:linear-chain}, 
 $\xn$ is the $\tn \times \dim$ matrix of feature descriptors corresponding to sequence $n$. 
\subsection{Expectation estimates}
In order to estimate the expectations in Equations \eqref{eq:estimates-full-begin} to 
\eqref{eq:estimates-full-end}, we use a simple Monte Carlo approach where we draw 
samples from our approximate distributions and compute the empirical expectations. 
For example, for the $\ellterm$ we have:
\begin{align}
\label{eq:ellhat}
\elltermhat = \frac{1}{S} \sum_{n=1}^{\nseq} \sum_{k=1}^K \pi_k \sum_{i=1}^{S}  \log p(\yn | \funnki, \fbinni) \text{,} 
\end{align}
\text{with }  $\funnki  \sim  \Normal( \qfmean{k(n)}, \qfcov{k(n)})$ and 
$\fbinni \sim \Normal(\mbin, \Sbin)$,
 for $i = 1, \dots, S$, where $S$ is the number of samples used, 
 and each of the individual blocks of  $\qfmean{k(n)}$ and $\qfcov{k(n)}$ are given 
 in Equation \eqref{eq:bkjn-skjn}. We use a similar approach for estimating the gradients 
 of the $\ellterm$ and they are given in the supplementary material.

\section{Learning}
We learn the parameters of our model, i.e. the  parameters of our approximate variational posterior  
well as the hyperparameters ($\{ \varparam, \hyper \}$) through gradient-based optimization of 
the variational objective ($\elbo$). One of the main advantages of our method is the decomposition 
of the $\ellterm$ in Equation \eqref{eq:ellhat} and its gradients as a sum of  expectations of the individual 
likelihood terms for each sequence. This result enables us to use parallel computation and 
stochastic optimization in order to make our algorithms useful in practice. 

Therefore, we consider batch optimization for small-scale problems (exploiting parallel computation) and 
stochastic optimization techniques for larger problems. Nevertheless, from a statistical perspective, learning 
in both settings is still hard due to the noise introduced by the empirical expectations (in both the 
batch and the stochastic setting) and the noisy gradients when using stochastic learning 
frameworks such as stochastic gradient descend (\sgd). In order to address these issues, we 
use variance reduction techniques such as control variates in the batch case. In the stochastic 
setting, in addition to standard control variates used in sampling methods and 
some stochastic variational frameworks \citep{ranganath-et-al-aistats-2014}, we use 
the recently developed \saga method for optimization. We describe 
in section \ref{sec:var-reduction} why these two
approaches, standard control variates and \saga, are complementary and should improve 
learning in our method. 

\paragraph{Computational complexity}
The time-complexity of our stochastic optimization is dominated by the
computation of the posterior's entropy, Gaussian sampling, and running the forward-backward
algorithm, which yields an overall
cost of $O(\m^3 + \tn^3 + S \tn|\V|^2)$. 
The space complexity is dominated by 
storing inducing-point covariances, 
which is $O(\m^2)$. To put this in the perspective of other available methods, 
the existing Bayesian structured model with 
\ess sampling \citep{bratieres-et-al-tpami-2015} has time and memory complexity of $O(N^3)$ and $O(N^2)$ respectively,  where $N$ is the total number of observations (e.g.~words). \crf's time and space complexity with stochastic optimization depends on the feature dimensionality, 
i.e.~it is $O(D)$. The actual running time of \crf  also depends on the cost of model selection via a cross-validation procedure. \ess sampling makes the method of  \citet{bratieres-et-al-tpami-2015} completely unfeasible for large datasets and \crf has high running times for problems with high dimensions  and many hyperparameters. Our work aims to make Bayesian structured prediction practical for large datasets, while being able to use 
infinite-dimensional feature spaces as well as
 sidestepping a costly cross-validation procedure.
\subsection{Variance reduction techniques
	\label{sec:var-reduction}}
Our goal is to approximate an expectation of a function $g(\f)$ 
over the random variable 
$\f$ that follows a distribution $q(\f)$, i.e. $\mathbb{E}_q[g(\f)]$ via Monte Carlo samples. 
The simplest way to reduce the variance of the empirical estimator $\bar{g}$ is to 
subtract from $g(\f)$ another function $h(\f)$ that is highly correlated with $g(\f)$. 
That is, the function $\tilde{g}(\f):= g(\f) - \hat{a} h(\f)$ will have the 
same expectation as $g(\f)$ i.e. $\mathbb{E}_q[\tilde{g}] = \mathbb{E}_q[g]$, 
provided that $\mathbb{E}_q[h] = 0$ \footnote{We note that, 
in general, to ensure unbiasedness, $\mathbb{E}_q[h]$, 
if easily and efficiently computable, can be subtracted from $h$ to form an estimator 
$\tilde{g}:= g - h + \mathbb{E}_q[h]$.}.  
More importantly, as the variance of the new function is 
$\text{Var}[\tilde{g}] = \text{Var}[g] + \hat{a}^2\text{Var}[h] - 2 \hat{a}\text{Cov}[g, h]$,
our problem boils down to finding suitable $\hat{a}$ and $h$ so as to minimize 
 $\text{Var}[\tilde{g}]$.
The following two techniques are based on this simple principle and their main 
difference lies upon the distribution over which we want to reduce the variance.
\paragraph{Standard control variates for reducing the variance w.r.t.~the variational distribution.} 
Here  $q(\f)$  is the variational distribution and $g(\f) =\grad_{\lambda} \log q(\f) \log p(\yn | \fn)$ 
(see supplementary material). Previous work \citep{ranganath-et-al-aistats-2014,dezfouli-bonilla-nips-2015} has found that a suitable correction term 
is given by $h(\f) =\grad_{\lambda} \log q(\f)$, which has expectation zero. Given this, the optimal $\hat{a}$ can
be computed as $\hat{a} = {\text{Cov}[g,h]}/{\text{Var}[h]}$. 
The use of control variates is essential 
for the effectiveness of our framework. for example, in our experiments
described in \S \ref{sec:experiments} we have found 
that, in the batch setting, their use reduces the error rate for the Japanese name entity recognition task from about 46\% to around 5\%.

\paragraph{SAGA for reducing the variance w.r.t.~the data distribution.}
The fast incremental gradient method (\saga) has been recently proposed as a better alternative to 
existing stochastic optimization algorithms. Here the $q$ distribution we want to reduce the variance over is 
the data distribution $p(\mat{X},\mat{Y})$; $g(\f)$ is the per-sample gradient direction;  
and $h(\f)$ is the past stored gradient direction at the same sample point. 
Since the expectation of the past stored gradient will be non-zero, 
\saga \citep{defazio2014saga} uses the general estimator $\tilde{g}(\f) := g(\f) - h(\f) + \mathbb{E}_q[h(\f)]$. 
The quantity $\mathbb{E}_q[h(\f)]$ is an average over past gradients. We note that, crucially 
for our model,   this average can be cached instead of re-calculated at each iteration.

\section{Experiments \label{sec:experiments}}
For comparison purposes, we used the same benchmark dataset suite as that used by \citet{bratieres-et-al-tpami-2015}, 
which targets several standard \nlp problems and is part of the \crf{++} toolbox\footnote{This was developed by  Taku Kudo  and 
can be found at \url{https://taku910.github.io/crfpp/}.}.
 This includes 
  noun phrase 
identification (\basenp); chunking, i.e.~shallow parsing labels sentence constituents (\chunking);
identification of word segments in sequences of Chinese ideograms (\segmentation); 
and Japanese named entity recognition (\japanesene). As we will see, 
on these tasks our approach is on par with 
competitive benchmarks which,
unlike our method, exploit the structure of the likelihood. 

For more details of these datasets and 
the experimental set-up for reproducibility of the results see the supplementary material.
\subsection{Small-scale experiments}
\begin{table}
\caption{Mean error rates and standard deviations in brackets on small-scale experiments using 
5-fold cross-validation. The average number of observed words ($N$) on these problems range from
942 to 3740. \svm corresponds to structured support vector machines; 
\crf to conditional random fields; \gpess corresponds to \gpstruct with \ess for inference \citep{bratieres-et-al-tpami-2015}; 
\gpvarb and \gpvars correspond to our method with batch optimization  and
stochastic optimization respectively; and \gppseudo corresponds to our method with stochastic optimization using a 
piecewise pseudo-likelihood.
\label{tab:error-small}
}
\centering
\begin{tabular}{l c c c c c c c c }
\toprule
Dataset &  \multicolumn{5}{c}{Method} \\
\midrule
    & \svm  & \crf & \gpess & \gpvarb & \gpvars & \gppseudo\\
\cmidrule{2-7}
    \basenp	& 5.91   (0.44) 	& 5.92  (0.23) 	& 4.81 (0.47) 	& 5.17	(0.41) 	& 5.27 (0.24) & 5.37 (0.33)\\
    \chunking	 	& 9.79  (0.97) 	& 8.29 (0.77) 	& 8.77 (1.08) 	& 8.76	(1.09) 	& 10.02 (0.41) & 9.58 (0.87)\\
    \segmentation 	& 16.21 (2.21) 	& 14.94 (5.65) 	& 14.88 (1.80)  & 15.61 (1.90) 	& 14.97 (1.38) & 15.16 (1.57)\\
    \japanesene 	& 5.64 (0.82) 	& 5.11 (0.66) 	& 5.83 (0.83) 	& 5.23	(0.68) 	& 4.99 (0.41) & 4.80 (0.65)\\

\bottomrule
\end{tabular}

\end{table}

Table \ref{tab:error-small} shows the error rates on the small 
experiments across the different datasets considered. Overall, 
we observe that our method in batch mode (\gpvarb) is consistently 
better than \svm and compares favorably with \crf. When compared 
to \gpess, both versions of our method, the batch and the stochastic, 
also have similar performance with the notable exception of \gpvars on
\chunking. However, we do note that \gpvars has the smallest standard deviation among all compared 
methods over all datasets. We credit this desirable property to the usage of doubly controlled variates (SAGA + standard control variates), as well as to the conservative learning rates chosen for these tests.
From these results we can conclude that, despite not knowing the details of 
the conditional likelihood, our method is very competitive with other methods
that exploit this knowledge and has similar performance to \gpess.
\subsubsection{Accelerating inference with 
	a piecewise pseudo-likelihood}
In order to demonstrate the flexibility of our approach, we also tested the performance of our framework 
when the true likelihood is approximated by a piecewise pseudo-likelihood \citep{SutMcC07} that only takes in consideration the
local interactions within a single factor between the variables in our model. 
We emphasize that this change did not require 
any modification to our inference engine 
and we simply used this pseudo-likelihood 
as a drop-in replacement for the exact 
likelihood. 
As we can see from the results in 
Table \ref{tab:error-small} (\gppseudo), 
the performance of our model under this regime is comparable to the one for \gpvars.
Furthermore,  
every step of stochastic optimization ran roughly twice as fast in \gppseudo as in \gpvars, which 
made up for the fact that for a linear-chain structure the computation time of forward-backward is quadratic in the label cardinality while for the piecewise pseudo-likelihood the cost is linear.
Such an approach might be considered for extending our 
framework to models such as grids or skip-chains, for which the evaluation of the true structured likelihood would be intractable.
  Alternatively, a structured mean field approximation using tractable approximating families of sub-graphs (linear chains, for instance) might be used for the same purpose. 
\subsection{Larger-scale experiments}
Here we report the results on an experiment that used the largest dataset in our 
benchmark suite (\basenp). For this dataset we used a five-fold cross-validation 
setting and $\nseq= 500$ training sequences. This amounts to 
roughly $11,611$ words on average. For testing we used the remaining ($323$) sequences.
In this setting \gpess is completely impractical. We compare the results of 
our model with \crf, which from our previous experiment was the most competitive 
baseline. 
Unlike the small experiments where the regularization parameter was learned through 
cross-validation, because of the large execution times, here we report the error rates 
for two values of this parameter $\lambda_\text{reg} \in \{0.1, 1\}$, where we obtained 
$5.13\%$ and  $4.50\%$ respectively. Our model (\gpvars) attained an error rate of 
$5.14\%$, which is comparable to  \crf's performance. As in the small experiments, we 
conclude that our model, despite not knowing the details of the likelihood, it performs  
on par with methods that were hard-coded for these types of likelihoods. 
See the supplementary material for more analysis. 

\section{Related work}
Recent advances in sparse \gp models for regression
 \citep{titsias-aistats-2009,hensmangaussian} have allowed the applicability  of such models 
 to very large datasets, opening opportunities for the  extension of these
 ideas  to classification and to problems with generic \iid likelihoods 
 \citep{hensman-et-al-aistats-2015, nguyen-bonilla-nips-2014, dezfouli-bonilla-nips-2015,hensman-et-al-nips-2015}. 
 However, none of these approaches is actually applicable to structured prediction problems, which inherently deal with  
 non-\iid likelihoods. 
 
Twin Gaussian processes \citep{bo2010twin} address 
structured continuous-output problems by forcing input kernels to be similar to output kernels. In contrast, here we deal 
with the harder problem of structured \emph{discrete}-output problems, where one usually requires computing  
expensive likelihoods during training.  
The structured 
continuous-output problem  is somewhat related to the area of multi-output regression with \gp{s}  
for which, unlike discrete structured prediction with \gp{s}, the literature is  relatively mature 
\citep{alvarez2010efficient,alvarez2011computationally,alvarez-lawrence-nips-08,bonilla-et-al-nips-08}.

The original structured Gaussian process model, \citep[\gpstruct, ][]{bratieres-et-al-tpami-2015} uses 
Markov Chain Monte Carlo (\mcmc) sampling as the inference method and is not equipped with sparsification techniques 
that are crucial for scaling to large data. 
 \citet{bratieres-et-al-icml-2014} have explored a distributed 
version of \gpstruct based on the pseudo-likelihood approximation \citep{Besag1975} where several weak learners are 
trained on subsets of \gpstruct's latent variables and bootstrap data. 
However, within each weak learner,  inference is still done via \mcmc. A variational alternative for \gpstruct inference \citep{Srijith-et-al-2014} is also available. 
However, it relies on pseudo-likelihood approximations and was only evaluated on small-scale problems. 
Unlike this work, our approach can deal with both pseudo-likelihoods and generic (linear-chain) structured likelihoods, 
and we rely on our sparse approximation procedure and our automated variational inference technique -- rather than on bootstrap aggregation -- to achieve good performance on larger datasets.

\section{Conclusion \& discussion}
We have presented a Bayesian structured prediction model with 
 \gp priors and linear-chain likelihoods. We have 
developed an automated variational inference algorithm that is statistically efficient in 
that only requires expectations over very low-dimensional Gaussians in order to 
estimate the expected likelihood term in the variational objective. We have exploited 
these types of theoretical insights as well as practical statistical and optimization 
tricks to make our inference framework scalable and effective.
Our model generalizes recent advances in 
\crf{s} \citep{koltun2011efficient} by allowing
general positive definite kernels defining their energy functions and opens new directions for combining 
deep learning with structure models \citep{zheng2015conditional}.

As mentioned in the introduction, for general structured prediction problems 
one may need to set up the configuration of 
the latent functions (e.g.~the 
unary and pairwise functions in the linear-chain case). Thus, the process of developing an inference procedure for a different structure (e.g.~when going from linear chains to skip-chains) requires  some human intervention.
Nevertheless, when applied to fixed structures 
our approach is entirely “black box” with respect to the choice of likelihood, inasmuch as different likelihoods can be used without any manual change to the inference engine.  

Furthermore, we have already seen in our small-scale experiments a possible 
way  to extend our method to more general structured likelihoods,
where the exact likelihood is replaced by 
a piecewise pseudo-likelihood.
Such an approach might be considered for using our  
framework in models such as grids or skip-chains, for which the evaluation of the true structured likelihood would be intractable.
The performance of our small-scale experiments in which the true likelihood was approximated by its pseudo-likelihood was very encouraging and
we leave a more 
in-depth investigation of the efficacy of this approach for future work.
We also leave to future work the challenging task of automating the very procedure that turns a structured  specification into a likelihood-agnostic inference procedure. 

Overall,  we believe our approach  is a fundamental step to developing automated inference methods 
for general structured prediction problems.

\section*{Supplementary Material}
\appendix
\section{Proof of Theorem 2}
Here we proof the result that we can estimate the expected log likelihood and its 
gradients using expectations over low-dimensional Gaussians.
\subsection{Estimation of $\ellterm$ in the full (non-sparse) model}
For the $\ellterm$ we have that: 
\begin{align}
\ellterm &= \Eb{\sum_{n=1}^\nseq \log p(\yn | \fn)}_{\qun \qbin} \\
&= \sum_{n=1}^{\nseq}  \int_{\fbin} \int_{\fun} \qun \qbin \log p(\yn | \fn) \  d\fun d\fbin \\
& =  \sum_{n=1}^{\nseq}  \int_{\fbin}  \int_{\funn} \int_{\funnotn} q(\funnotn | \funn) q(\funn) \qbin 
\log p(\yn | \fn) \ d\funnotn d\funn d\fbin  \\
&= \sum_{n=1}^{\nseq} \Eb{\log p(\yn | \fn)}_{q(\funn) \qbin } \\
\label{eq:ell-full-end}
& = \sum_{n=1}^{\nseq} \sum_{k=1}^{K} \pi_k  \Eb{\log p(\yn | \fn)}_{\qkunn \qbin } \text{,}
\end{align}
where $ \qkunn$ is a $(\tn \times \Vsize)$-dimensional Gaussian with block-diagonal covariance $ \qfcov{k(n)}$, each 
block of size $\tn \times \tn$. Therefore, we can estimate the above term by sampling from 
$\tn$-dimensional Gaussians independently. Furthermore, $\qbin$ is a $\Vsize^2$-dimensional Gaussian, which 
can also be sampled independently. In practice, we can 
assume that the covariance of $\qbin$ is diagonal and we only sample from unary Gaussians for the 
pairwise functions.
\QEDA
\subsection{Gradients \label{sec:proof-grad-full}}
Taking the gradients of the k$\mth$ term for the n$\mth$ sequence in the $\ellterm$:
\begin{align}
  \elltermkn &=   \Eb{\log p(\yn | \fn)}_{\qkunn \qbin } \\
       & = \int_{\fbin} \int_{\funn} \qkunn \qbin  \log p(\yn | \fn) \ \dfunn \dfbin \\
\grad_{\lambdaku}  \elltermkn &=   \int_{\fbin} \int_{\funn} \qkunn \qbin \grad_{\lambdaku} \log \qkunn  \log p(\yn | \fn) \ \dfunn \dfbin \\
\label{eq:grad-unary-full}
& = \Eb{\grad_{\lambdaku} \log \qkunn \log p(\yn | \fn)}_{\qkunn \qbin } \text{,}
\end{align}
where we have used the fact that $\nabla_{\vec{x}} f(\vec{x}) = f(\vec{x}) \nabla_{\vec{x}} \log f(\vec{x})$ for any nonnegative function 
$f(\vec{x})$
Similarly.  the gradients of the parameters of the distribution over binary functions can be estimated using:
\begin{equation}
  \grad_{\lambdabin}  \elltermkn = \Eb{\grad_{\lambdabin} \log \qbin \log p(\yn | \fn)}_{\qkunn \qbin } \text{.}
\end{equation}
\QEDA
\section{KL terms in the sparse model}
The KL term ($\klterm$) in the variational objective ($\elbo$) is composed of a KL divergence between 
the approximate posteriors and the priors over the inducing variables  and pairwise functions:
\begin{equation}
\klterm =  \underbrace{ - \kl{q(\u)}{p(\u)} }_{\klun}  \underbrace{ - \kl{q(\fbin)}{p(\fbin)}  }_{\klbin}\text{,} 
\end{equation}
where, as the approximate posterior and the prior over the pairwise functions are Gaussian, the KL 
over pairwise functions can be computed analytically:
\begin{align}
  \klbin = - \kl{q(\fbin)}{p(\fbin)}  &= \kl{\Normal(\fbin; \mbin, \Sbin)}{\Normal(\fbin; \vec{0}, \Kbin)} \\
   &=
  - \frac{1}{2} \left( \log \det{\Kbin}  - \log \det{\Sbin} 
   +  \mbin^T \Kbin^{-1} \mbin  + \trace{\Kbin^{-1} \Sbin} - \Vsize \right) \text{.}
\end{align}
For the distributions over the unary functions we need to compute a KL divergence between 
a mixture of Gaussians and a Gaussian.
For this we consider the decomposition of the KL divergence as follows:
\begin{equation}
  \klun = - \kl{\qu}{p(\u)} = \underbrace{\expectation_q[- \log \qu]}_{\enterm} + \underbrace{\expectation_q[\log p(\u)]}_{\crossterm} \text{,}
\end{equation}
where the entropy term ($\enterm$) can be lower bounded using Jensen's inequality:
\begin{equation}
  \label{eq:entropy}
  \enterm \geq - \sum_{k=1}^{\k} \pi_k \log \sum_{\ell=1}^\k \pi_{\ell} \Normal(\postmean{k}; \postmean{\ell}, \postcov{k} + \postcov{\ell}) 
  \defeq \entermhat \text{.}
\end{equation}
and the negative cross-entropy term ($\crossterm$) can be computed exactly:
\begin{equation}
  \crossterm = - \frac{1}{2} \sum_{k=1}^\k \pi_k \sum_{j=1}^{\Vsize} [\m \log 2 \pi + \log \det{\Kzz} 
  + \postmean{kj}^T \Kzzinv \postmean{kj} + \trace{\Kzzinv \postcov{kj}}] \text{.}
\end{equation}
\section{Proof of Theorem 3}
To prove Theorem 3 we will express the expected log likelihood term in the same form as that 
given in Equation \eqref{eq:ell-full-end}, showing that the resulting 
$\qkunn$ is also a  $(\tn \times \Vsize)$-dimensional Gaussian with block-diagonal covariance,  
having $\Vsize$ blocks each of dimensions $\tn \times \tn$.
We start by taking the given $\ellterm$, where the expectations are 
over the joint posterior $q(\f, \u | \llambda) = p(\fun | \u) q(\u) q(\fbin)$:
\begin{align}
\ellterm &= \Eb{\sum_{n=1}^\nseq \log p(\yn | \fn)}_{p(\fun | \u) q(\u) q(\fbin) } \\
& = \int_\f \log p(\y | \f)  \underbrace{\int_\u q(\u) p(\fun | \u) \du }_{q(\fun)}q(\fbin)\df \text{,}
\end{align}
where our our approximating distribution is:
\begin{align}
q(\f) &= q(\fun) q(\fbin) \\
q(\fun) & =  \int_\u q(\u) p(\fun | \u) \du \text{,}
\end{align}
which can be computed analytically:
\begin{align} 
\label{eq:qfunsparse}
q(\fun) &= \sum_{k=1}^{K} \pi_k q_k(\fun) =  \sum_{k=1}^{K} \pi_k \prod_{j=1}^{\Vsize} \Normal(\funj; \qfmean{kj}, \qfcov{kj}) \\
\label{eq:meanqfsparse}
\qfmean{kj} & = \Aj \postmean{kj} \\
\label{eq:covqfsparse}
\qfcov{kj} &= \priorcov + \Aj \postcov{kj} \Aj^T \text{.}
\end{align}
We note in Equation \eqref{eq:qfunsparse} that  $q_k(\fun)$ has a block diagonal structure, which implies that 
we have the same expression for the $\ellterm$ as in Equation  \eqref{eq:ell-full-end}. Therefore, we obtain 
analogous estimates:
\begin{align}
\ellterm &= \sum_{n=1}^{\nseq} \sum_{k=1}^{K} \pi_k  \Eb{\log p(\yn | \fn)}_{\qkunn \qbin } \text{,}
\end{align}
Here, as before,  $\qkunn$ is a $(\tn \times \Vsize)$--dimensional Gaussian with block-diagonal covariance $ \qfcov{k(n)}$, each 
block of size $\tn \times \tn$. The main difference in this (sparse) case is that 
$\qfmean{k(n)}$ and $\qfcov{k(n)}$ are constrained by the expressions in Equations 
\eqref{eq:meanqfsparse} and \eqref{eq:covqfsparse}.
Hence, the proof for the gradients follows the same derivation as 
in \S \ref{sec:proof-grad-full} above.
\QEDA
\section{Gradients of $\elbo$ for sparse model}
Here we give the gradients of the variational objective wrt the parameters for the variational
distributions over the inducing variables, pairwise functions and hyper-parameters.
\subsection{Inducing variables}
\subsubsection{KL term}
As the structured likelihood does not affect the KL divergence term, 
the gradients corresponding to this term are similar to those in the non-structured case 
\citep{dezfouli-bonilla-nips-2015}.  
Let $\Kzzall$ be the block-diagonal covariance with $\Vsize$ blocks $\Kzz$, $j=1, \ldots Q$.  Additionally,
lets assume the following definitions:
\begin{align}
  \label{eq:Ckl}
  \Ckl  & \defeq \postcov{k} + \postcov{\ell} \text{,}\\
  \nkl  & \defeq \Normal(\postmean{k}; \postmean{\ell}, \Ckl) \text{,} \\
  z_k  &\defeq \sum_{\ell=1}^\k \pi_{\ell} \nkl \text{.}
\end{align}

The gradients of 
 $\klterm$ wrt the posterior mean and posterior covariance for component $k$ are: 
\begin{align}
\grad_{\postmean{k}} \crossterm &= - \pi_k \Kzzallinv \postmean{k}  \text{,}\\
\grad_{\postcov{k}} \crossterm &= - \frac{1}{2} \pi_k \Kzzallinv  \\
\grad_{\pi_k} \crossterm &= - \frac{1}{2}  \sum_{j=1}^{\Vsize} [\m \log 2 \pi + \log \det{\Kzz} 
  + \postmean{kj}^T \Kzzinv \postmean{kj} + \trace{\Kzzinv \postcov{kj}}] 
\text{,}
\end{align}
where we note that  we compute $\Kzzallinv$ by inverting the corresponding blocks $\Kzz$ independently. 
The gradients of the entropy term wrt the variational parameters are:
\begin{align}
  \label{eq:grad-ent-init}
  \grad_{\postmean{k}} \entermhat &= \pi_k \sum_{\ell=1}^\k \pi_{\ell} \left( \frac{\nkl}{\zk} + \frac{\nkl}{\zl} \right)   \Ckl^{-1} (\postmean{k} - \postmean{\ell}) \text{,} \\
  \grad_{\postcov{k}} \entermhat &= \frac{1}{2} \pi_k \sum_{\ell=1}^{\k} \pil \left(  \frac{\nkl}{\zk} + \frac{\nkl}{\zl} \right) 
  \left[ \Ckl^{-1} - \Ckl^{-1}  (\postmean{k} - \postmean{\ell})  (\postmean{k} - \postmean{\ell})^T \Ckl^{-1}  \right] \text{,} \\
  \nonumber
  \label{eq:grad-ent-end}
  \grad_{\pi_k} \entermhat &= - \log \zk - \sum_{\ell=1}^{\k} \pi_{\ell} \frac{\nkl}{\zl} \text{.}
\end{align}
\subsubsection{Expected log likelihood term}
Retaking the gradients in the full model In Equations \eqref{eq:grad-unary-full}, we have that:
\begin{align}
  \grad_{\lambdaku} \elltermkn & = \Eb{\grad_{\lambdaku} \log \qkunn \log p(\yn | \fn)}_{\qkunn \qbin } \text{,}
\end{align}
where the variational parameters  $\lambdaku$ are the posterior means and covariances 
($\{ \postmean{kj} \}$ and $\{ \postcov{kj} \}$)
of 
the inducing variables. As given in Equation \eqref{eq:qfunsparse}, $\qkun$ factorizes  over the latent process
($j=1, \ldots, \Vsize$), so do the marginals $\qkunn$, hence:
\begin{align}
  \label{eq:gradlogqkun}
  \grad_{\lambdaku} \log \qkunn =  \grad_{\lambdaku} \sum_{j=1}^{\Vsize} \log \Normal(\funnj; \bkjn, \Sigmakjn) \text{,}
\end{align}
where each of the distributions in Equation \eqref{eq:gradlogqkun} is a $\tn$--dimensional Gaussian.
Let us assume the following definitions:
\begin{align}
  \Xn &: \text{all feature vectors corresponding to sequence }n\\ 
  \Ajn &\defeq \knz \Kzzinv \\
  \tildeKjn &\defeq \knn - \Ajn \kzn \text{, therefore:} \\
  \bkjn &= \Ajn \mkj \text{,}\\
  \Sigmakjn & = \tildeKjn + \Ajn \Skj \Ajn^T \text{.}
\end{align}
Hence, the gradients of $\log \qkun$ wrt the the variational parameters of the unary posterior distributions over
the inducing points are:
\begin{align}
  \grad_{ \postmean{kj}} \log \qkunn & = \Ajn^T \Sigmakjn^{-1} \left( \funnj - \bkjn \right) \text{,} \\
  \grad_{ \postcov{kj}} \log \qkunn &= \frac{1}{2} 
   \Ajn^T \left[ 
     \Sigmakjn^{-1} (\funnj - \bkjn) (\funnj - \bkjn)^T \Sigmakjn^{-1} - \Sigmakjn^{-1} 
     \right] \Ajn 
\end{align}
Therefore, the gradients of $\ellterm$ wrt the parameters of the distributions over unary functions are:
\begin{align}
\grad_{\postmean{kj}} \ellterm & = \frac{\pi_k}{S} \Kzzinv \sum_{n=1}^{\nseq} \kzn \Sigmakjn^{-1} 
    \sum_{i=1}^S (\funnj^{(k,i)}  - \bkjn ) \log p(\yn | \funnki , \fbinni) \text{,} \\
\grad_{\postcov{kj}} \ellterm & = \frac{\pi_k}{2 S} \sum_{n=1}^{\nseq} \Ajn^T 
  \Big\{
    \sum_{i=1}^S \big[ \Sigmakjn^{-1} (\funnj^{(k,i)} - \bkjn)(\funnj^{(k,i)} - \bkjn)^T \Sigmakjn^{-1}  \\
    \nonumber
    & \quad\qquad\qquad\qquad\qquad - \Sigmakjn^{-1}\big] \log   p(\yn | \funnki , \fbinni) 
  \Big\}    
  \Ajn
\end{align}
\subsubsection{Pairwise functions}
The gradients of the $\klbin$ wrt the parameters of the posterior over pairwise functions  are given by:
\begin{align}
\grad_{\mbin} \klbin &= - \Kbin^{-1} \mbin \\
\grad_{\Sbin} \klbin &= \frac{1}{2} \left( \Sbin^{-1} - \Kbin^{-1}  \right)
\end{align}
The gradients of the $\ellterm$ wrt the parameters of the posterior over pairwise functions  are given by:
\begin{align}
  \grad_{\mbin}\ellterm &= \frac{1}{S} \sum_{n=1}^\nseq \sum_{k=1}^K \pi_k \sum_{i=1}^S
  \Sbin^{-1}(\fbinni - \mbin)  \log   p(\yn | \funnki , \fbinni) \\
  \grad_{\Sbin}\ellterm &= \frac{1}{2S} \sum_{n=1}^\nseq \sum_{k=1}^K \pi_k \sum_{i=1}^S
  [\Sbin^{-1}(\fbinni - \mbin)(\fbinni - \mbin)^T \Sbin^{-1} - \Sbin^{-1} ]  \log   p(\yn | \funnki , \fbinni) 
\end{align}
\section{Experiments}
\subsection{Experimental set-up}
Details of the benchmarks used in our experiments can be seen in Table \ref{tab:datasets}.
\begin{table}
\centering
\caption{Datasets used in our experiments. \label{tab:datasets}. For each dataset we see the 
number of categories (or vocabulary $\Vsize$), the number of features($\dim$),  the number of 
training sequences used in the small experiments ($\nseq$ small), and the average (across folds)
number of training words for the small experiments ($\bar{\n}$). 
}
\begin{tabular}{ccccc}
Dataset & $\Vsize$ & $\dim$ & $\nseq$ small & $\bar{\n}$ small \\
\toprule
\basenp   &3    &6,438    &150  & 3739.8\\
\chunking   &14   &29,764   &50     & 1155.8\\
\segmentation &2    &1,386    &20   & 942 \\  
\japanesene &17   &102,799    &50   & 1315.4\\  
\bottomrule
\end{tabular}
\end{table}
For the experiments with batch optimization, we optimized the three sets of parameters separately in a global loop (variational parameters for unary nodes, variational parameters for pairwise nodes, and hyper-parameters). In each global iteration, each set of parameters were optimized while keeping the rest of the parameters fixed. Variational parameters for unary nodes were optimized for 50 iterations, variational parameters for pairwise nodes were optimized for 10 iterations, and hyper-parameters were updated for 5 iterations. We used L-BFGS algorithm for optimizing each set of parameters, and parameters were optimized for a maximum of 5 1/2 hours, or until the convergence, whichever comes first. Convergence was detected when the objective function in two consecutive global iterations was less than 1e-05, or the average change in the variational parameters for unary nodes was less than 0.001. The reported results are the predictions based on the best objective function achieved during the optimization. 10,000 samples ($S=10,000$) were used for approximating expected log likelihood and its gradients and  $10\%$ of these samples were 
used for the optimal $\hat{a}$ in the control variate calculation. For all the experiments 500 inducing points were used ($M=500$).

In experiments with stochastic optimization, similar to the experiments with batch optimization, each set of parameters were optimized separately. In each global iteration, variational parameters for unary nodes were updated for 3000 iterations and variational parameters for pairwise nodes were updated for 1000 iterations (hyper-parameters were not optimized in the stochastic optimization experiments, and they were fixed to 1). 4,000 samples were used for estimating expected log likelihood and its gradients ($S=4000$). Similar to the batch optimization case, we used 500 inducing points ($M=500$). The step-size for updating the means of the inducing points was set to 1e-4, and the step-size for updating the covariances of the inducing points were set to 1e-5.
\subsection{Performance profiles}
\begin{figure}
\includegraphics[width=0.5\textwidth]{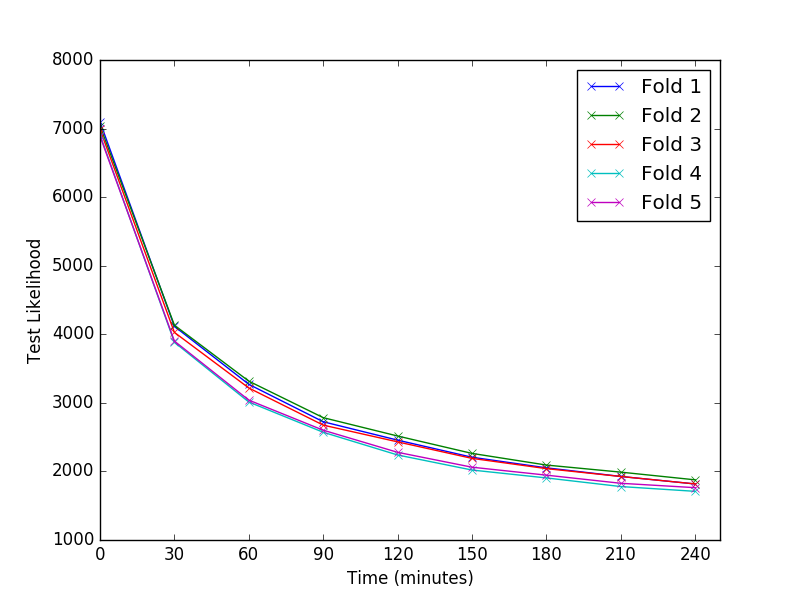}
\includegraphics[width=0.5\textwidth]{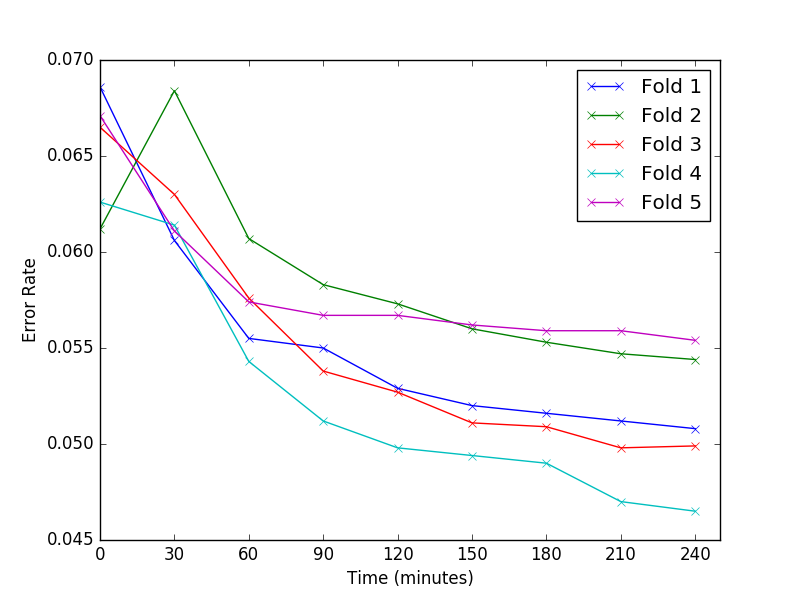}
\caption{The test performance of \gpvars on \basenp for the large scale experiment as a function of time. 
\label{fig:-perf-profile}
}
\end{figure}

Figure \ref{fig:-perf-profile} shows the performance of our algorithm as a function of time.
We see that the test likelihood decreases very regularly in all the folds and so does overall the error rate, albeit with more variability. 
The bulk of the optimization, both with respect to the test likelihood and with respect to the error rate, occurs during the first 120 minutes. This suggests that the kind of approach described in this paper might be particularly suited for cases in which speed of convergence is a priority.

{
\small
\bibliography{arxiv_graybox}
\bibliographystyle{plainnat}
}

\end{document}